\documentclass[sigconf, nonacm]{acmart}

\usepackage{pifont}
\usepackage{threeparttable}
\usepackage{booktabs}
\usepackage{multirow}
\usepackage{array, makecell}
\usepackage{utfsym}
\usepackage{algorithm}
\usepackage[noend]{algpseudocode}
\usepackage{graphicx}
\usepackage{textcomp}
\usepackage{xcolor}
\usepackage{natbib}
            
\begin{document}
\title{cuPilot: A Strategy-Coordinated Multi-agent Framework for CUDA Kernel Evolution\\
}


\author{Jinwu Chen}
\authornote{These authors contributed equally to this work.}
\affiliation{%
    \institution{Southeast University}
    \country{Nanjing, China}
}

\author{Qidie Wu}
\authornotemark[1]
\affiliation{%
    \institution{Tsinghua University}
    \country{Beijing, China}
}

\author{Bin Li}
\affiliation{%
    \institution{Tsing Micro}
    \country{Beijing, China}
}

\author{Lin Ma}
\affiliation{%
    \institution{Tsing Micro}
    \country{Beijing, China}
}

\author{Xin Si}
\affiliation{%
    \institution{Southeast University}
    \country{Nanjing, China}
}

\author{Yang Hu}
\affiliation{%
    \institution{Tsinghua University}
    \country{Beijing, China}
}

\author{Shouyi Yin}
\authornote{Corresponding authors.}
\affiliation{%
    \institution{Tsinghua University}
    \country{Beijing, China}
}
\email{yinsy@tsinghua.edu.cn}

\author{Jun Yang}
\authornotemark[2]
\affiliation{%
    \institution{National Center of Technology Innovation for EDA}
    \country{Nanjing, China}
}
\email{dragon@seu.edu.cn}

\begin{abstract}
Optimizing CUDA kernels is a challenging and labor-intensive task, given the need for hardware-software co-design expertise and the proprietary nature of high-performance kernel libraries.
While recent large language models (LLMs) combined with evolutionary algorithms show promise in automatic kernel optimization, existing approaches often fall short in performance due to their suboptimal agent designs and mismatched evolution representations. 
This work identifies these mismatches and proposes cuPilot, a strategy-coordinated multi-agent framework that introduces strategy as an intermediate semantic representation for kernel evolution. 
Key contributions include a strategy-coordinated evolution algorithm, roofline-guided prompting, and strategy-level population initialization.
Experimental results show that the generated kernels by cuPilot achieve an average speed up of 3.09$\times$ over PyTorch on a benchmark of 100 kernels. 
On the GEMM tasks, cuPilot showcases sophisticated optimizations and achieves high utilization of critical hardware units.
The generated kernels are open-sourced at https://github.com/champloo2878/cuPilot-Kernels.git.
\end{abstract}


\maketitle

\section{Introduction}
The optimization of CUDA kernels, a widely adopted GPU programming model~\cite{cuda}, is crucial for achieving peak AI inference/training performance.
However, optimizing GPU kernels is a labor-intensive and time-consuming task. 
First, it necessitates a thorough understanding of the target GPU architecture, along with meticulous tuning of both computation and memory access. It entails techniques like warp scheduling, fine-grained data layouts, and even PTX-level optimizations. 
Second, the scarcity of open-source kernels further elevates the barrier to high-performance kernel development, which compels designers to devote extensive time to exploring diverse optimization strategies. 
Moreover, the rapid iterative modern GPUs and applications inevitably render kernel optimization a challenging and inherently labor-intensive task. 

Fortunately, recent emerging LLMs have shown promise in automatic kernel generation. Many efforts have attempted to enhance LLM through reflection~\cite{gpuscentist}, Retrieval-Augmented Generation (RAG)~\cite{sakanaAI}, or Reinforcement Learning (RL)~\cite{kevin, cudal1}. 
However, most existing works remain focused on functional correctness, with significantly lower performance than manually written libraries. 
This limitation stems from their limited agentic flows, which merely rely on one-shot queries or inefficient reflections. 
For better exploitation of LLMs, integrating evolutionary algorithms with LLM agents is considered a viable path toward higher-order intelligence~\cite{alphaevolve, alphacode, lear}.
For GPU kernel, several works~\cite{sakanaAI, cudallm} explore crossover prompting techniques to generate new kernels from selected parent kernels, thus evolving kernels' performance across multiple generations.

However, kernels optimized by existing frameworks still remain suboptimal~\cite{sakanaAI_hf_datasets}, exhibiting few basic optimization strategies.
We identify that current kernel evolution frameworks suffer from multi-dimensional mismatches between the high-level evolutionary algorithm and low-level kernel code:
\textbf{(1) Mismatched crossover representation:} Conventional crossover prompting is operated at the code level, where optimization strategies are implicitly crossed. Such kernel-to-kernel prompting makes LLM to traverse an extended reasoning chain spanning strategy identification, strategy combination, and kernel synthesis. This often yields ineffective optimizations and drops previously attained gains, especially as the kernel's complexity grows.
\textbf{(2) Mismatched fitness representation:} The fitness function, based solely on performance, exhibits weak semantic correlation with the kernel code. This prevents LLM from accurately pinpointing bottlenecks among numerous profiling reports, thus resulting in trivial optimizations. 
\textbf{(3) Mismatched population initialization:} The initial population is represented by a sparse collection of kernel codes. It provides inadequate coverage of the entire optimization strategy space. Consequently, the optimization is prone to premature convergence to local optima.

To address these challenges, this work introduces \textbf{Strategy} as an intermediate representation and proposes a \textbf{Strategy-Coordinated} multi-agent framework named cuPilot. Key contributions are:
\begin{itemize}
\item A Strategy-Coordinated Evolution (SCE) algorithm is proposed, which decouples the long reasoning chain of conventional crossover prompting into strategy-level crossover and strategy-to-kernel translation. This enables LLM to optimize under explicit strategy guidance, substantially improving the viability and performance of generated kernels.

\item A roofline-guided prompting method is applied. By first positioning the target kernel on the GPU's roofline model, the LLM determines if it is compute-bound, memory-bound, or in a middle-zone, thus guiding the prompting for better strategy generation and kernel refinement.

\item A strategy-level population initialization method is proposed, which establishes an external strategy pool and learn optimization strategies from the historical data via RAG. It enhances the LLM's strategy generation capability and achieves a higher rate of evolutionary convergence.

\item Experiment on the 100 kernels of KernelBench~\cite{kernelbench}, cuPilot achieves $3.09\times$ speedup over PyTorch. The case study on GEMM tasks showcases sophisticated optimizations with high utilization of critical hardware units. The generated high-performance kernels are open-sourced.

\end{itemize}

\section{Background}

\subsection{CUDA Programming}

To achieve general-purpose GPU programming, NVIDIA developed CUDA (Compute Unified Device Architecture) as a parallel computing platform and programming model~\cite{cuda}. CUDA extended the C/C++ language to allow developers to leverage the massive parallel computing capabilities of GPUs. CUDA framework unlocks sophisticated control over GPU threads, memory hierarchies, and synchronization. Its programming abstraction also organizes threads into a hierarchical structure of warps, blocks, and grids. This capability renders it indispensable for accelerating computationally intensive tasks involving large-scale matrix and tensor operations~\cite{programming}. Furthermore, CUDA's ecosystem offers highly optimized libraries like cuBLAS and cuDNN~\cite{cudnn}, which significantly enhance application performance for linear algebra and deep neural networks. Therefore, the optimization of CUDA kernels is pivotal for effectively addressing the demands of complex modern applications.

\subsection{LLM-Assisted CUDA Kernel Generation}
Hand-optimization of CUDA kernels is a challenging and time-consuming task that requires designers to possess deep, multifaceted expertise in algorithms and hardware architecture. During the optimization process, designers rely heavily on extensive feedback to guide subsequent manual adjustments. 
This includes compiler feedback, reports from profiling tools, details of different architectures and instruction sets. 
These effort-intensive tasks significantly hinder the efficiency of kernel development, a challenge that is particularly acute given the rapid iteration of modern GPUs and applications. An example is FlashAttention kernel, whose adaptation to NVIDIA's Hopper architecture spanned nearly five years~\cite{flashattention}. Therefore, there is a pressing need for automated kernel development to reduce migration costs.

Advances in code-generation LLMs~\cite{dac-llmgen-1,dac-llmgen-2,dac-llmgen-3,dac-llmgen-4,dac-llmgen-5,dac-llmgen-6} have led researchers to leverage them for accelerating GPU kernel development. Some works have sought to improve LLM-generated kernels by integrating reinforcement learning (RL) frameworks~\cite{kevin,cudal1}, training dedicated models~\cite{autotriton}, and focusing the optimization objective on specific performance metrics (e.g., SwizzlePerf~\cite{swizzleperf} utilized L2 cache hit rate to guide LLM). Beyond these efforts, a common and viable strategy employs an evolutionary framework based on multi-agent systems. For instance, CUDA-LLM~\cite{cudallm} introduces the FSR framework, which guides LLMs to produce hardware-aware, high-performance CUDA kernels through the joint optimization of compilation and runtime performance. 
Furthermore, AI CUDA Engineer~\cite{sakanaAI} incorporates genetic algorithms into the LLM systems. It treats each kernel as an individual candidate and performs crossover prompting across generations of evolution, making a viable path for deep optimization of CUDA kernels.

\begin{figure}[!t]
\centering
\includegraphics[width=3.3in]{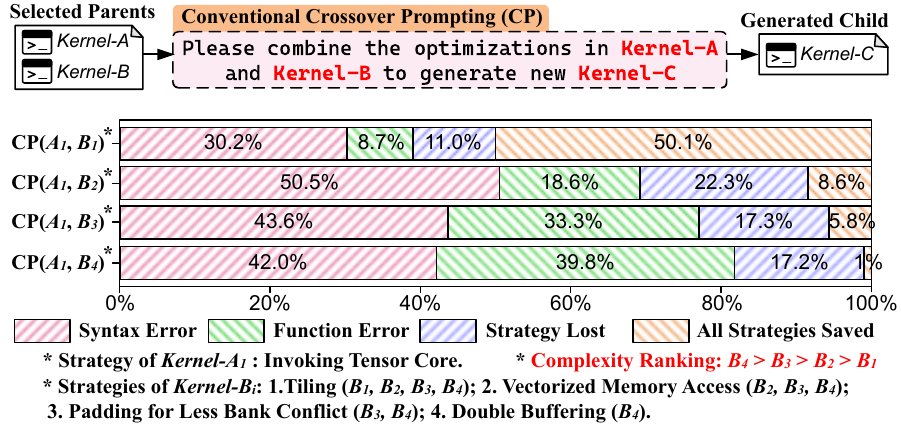}
\Description{nothing}
\caption{The impact of parent kernels' complexity on generated child kernels during conventional crossover prompting.}\label{crossover}
\end{figure}

\section{Motivation}

While existing works focus on enhancing performance through evolutionary frameworks, they nevertheless suffer from a significant oversight: 
the multi-dimensional mismatches between the high-level evolutionary algorithm and the low-level kernel code. 
These mismatches manifest in three dimensions: crossover representation, fitness representation, and population initialization.

\begin{figure*}[!t]
\centering
\includegraphics[width=7in]{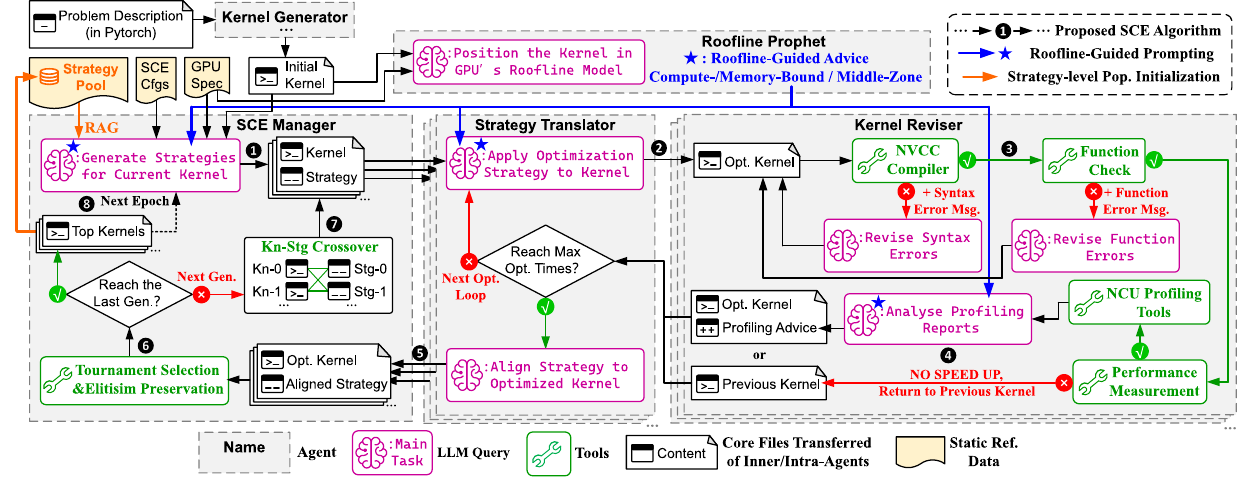}
\Description{nothing}
\caption{Overview of cuPilot multi-agent framework. Three key contributions are illustrated: SCE algorithm, roofline-guided prompting, and strategy-level population initialization.}\label{framework}
\end{figure*}

Firstly, conventional crossover prompting is reproduced on the Deepseek-R1, a SOTA reasoning model~\cite{deepseek-r1}, and the generated kernels are aggregated. 
As shown in the Figure~\ref{crossover}, the conventional crossover prompting requires the LLM to fuse two kernels directly, necessitating a long reasoning chain spanning strategy identification, strategy combination and kernel synthesis. This results in deteriorating kernel performance as the complexity increases. 
The crossover of $A_1$ and $B_1$ benefits from the low complexity of the source kernels. However, as the complexity of kernel B increases, the problem of representation mismatch becomes apparent. 
With complex strategies integrated into kernel B, the LLM struggles to accurately distinguish each strategy due to the semantic gap between strategies and codes. 
This difficulty leads to a significant degradation of results. 
As the figure shows, when the LLM performs crossover with the $B_2$ or $B_3$ kernels, a majority of the generated kernels contain syntax or function errors. 
Only 8.6\% and 5.8\%, respectively, successfully retain and combine all the original strategies from parents. Moreover, this success rate drops to a mere 1\% for the most complex kernel $B_4$. 
These results demonstrate that conventional crossover prompting, by working solely on low-level codes, is hard to match the crossover with kernel coding. 

Furthermore, mismatched fitness representation and population initialization pose critical challenges. The integration of evolutionary algorithms necessitates an appropriate fitness function for LLMs to evaluate generated kernels. 
Existing approaches primarily rely on overall kernel performance as the sole criterion. However, performance is inherently a high-dimensional metric wherein the actual bottlenecks remain implicit. Consequently, the model requires more granular and specific guidance to evolve effectively. 
Similarly, the scarcity of kernel data leads to poor population initialization. This results in an initial population with limited diversity of optimization strategies, which consequently restricts the optimization space and compromises the final quality.
To solve these issues, proposed cuPilot firstly introduces the SCE algorithm to mitigate the crossover representation mismatch. Then it utilizes the roofline model to guide LLMs prompting, and establishes an external strategy pool to improve the strategy-level population initialization.

\section{cuPilot Framework}
\subsection{Overview of cuPilot}

For better generating complex kernels and exploring the boundaries of existing LLMs, we propose cuPilot, a strategy-coordinated multi-agent evolutionary framework. Our core philosophy is to decompose the intricate task of kernel evolution into multiple simpler subtasks, enabling each agent to specialize effectively. 
Furthermore, for each subtask, we perform parallel queries to the LLM multiple times, which in effect boosts the sample population. This approach enables a thorough traversal of the optimization, with the best outcomes selected for subsequent subtasks, thereby improving the overall execution success rate.

\algnewcommand{\Input}{\item[\textbf{Input:}]}
\algnewcommand{\Inputx}{\item[\textbf{}]}
\algnewcommand{\Output}{\item[\textbf{Output:}]}
\begin{algorithm}[t!]
\caption{Strategy-Coordinated Evolution Algorithm}\label{SCE-Algorithm}
\begin{algorithmic}[1]
\Input Initial $K^{\text{init}}$,\;Epochs $E$,\;Generations $G$,\;
\Inputx Evolving Strategy Number List $[NS_0, NS_1, ..., NS_G]$,\;
\Inputx Evolving Kernel Number List $[NK_1, NK_2, ..., NK_G].$\;
\Output Optimized Kernel $K^{\text{opt}}.$

\For {$e = 1$ \textbf{to} $E$} 
    \State $S \gets \textbf{StrategyGenerator}(K^{\text{init}}, NS_0);$\;
    \State $P \gets \textbf{PopulationInitialization}(K^{\text{init}}, S);$\;
    \For {$g = 1$ \textbf{to} $G$}
        \For {$i = 1$ \textbf{to} $NS_{g-1}$}
            \State $(K_i', S_i) \gets \textbf{StrategyApplication}(P_i);$\;
            \State $P_i \gets \textbf{StrategyAlignment}((K_i', S_i));$\;
        \EndFor
        
        \State $K^g, S^g \gets \textbf{TournamentElitisim}(P, NK_g,NS_g);$\;
        \For {$i = 1$ \textbf{to} $NK_g$}
            \For {$j = 1$ \textbf{to} $NS_g$}
                \State $P \gets P \cup \{(K_i^g, S_j^g)\};$\;
            \EndFor
        \EndFor 
    \EndFor
    \State $K^{\text{init}} \gets \textbf{SelectTopKernels}(P);$\;
\EndFor
\State \textbf{return} $K^{\text{opt}} \gets \textbf{SelectBestKernel}(K^{\text{init}});$\;
\end{algorithmic}
\end{algorithm}

Based on this design principle, as illustrated in Figure~\ref{framework}, we partition the evolution process into three major functional regions, each corresponding to a dedicated agent: the SCE Manager, Strategy Translator, and Kernel Revisor. 
The SCE Manager operates at the \textbf{high level} to oversee the evolutionary population, handling the initialization and crossover of strategies. 
The Kernel Revisor functions at the \textbf{low level} to refine LLM-generated kernels, including corrections for syntax and function, and kernel profiling. 
cuPilot bridges the semantic gap between the evolutionary framework and the kernel code by introducing the Strategy Translator as an intermediary. 
Specifically, the Strategy Translator performs the following tasks: \textbf{(1) Applying an individual's strategy to its kernel and forwarding the kernel to the Kernel Revisor;} and \textbf{(2) Aligning the strategy with the revised kernel before passing the individual to the SCE Manager.} 
As a mediator, the Strategy Translator incorporates the strategy as an intermediate semantic abstraction, facilitating a handshake mechanism between the SCE Manager and Kernel Revisors. This ensures that the kernel evolution task is more comprehensible and executable by the LLM.

Additionally, we introduce the Roofline Prophet agent atop the overall SCE flow. 
This agent determines the kernel's position within the GPU's roofline model based on the kernel description and GPU specifications, subsequently providing optimization guidance to the three evolutionary agents. 
Moreover, we maintain a kernel strategy pool enriched with external strategy knowledge and historical evolving data, employing RAG to enhance the agent's strategy generation capability. 
For illustration brevity, we omit details on the Kernel Generator agent, which produces functionally correct vanilla kernels and is relatively straightforward to implement.

\begin{figure}[!t]
\centering
\includegraphics[width=3.3in]{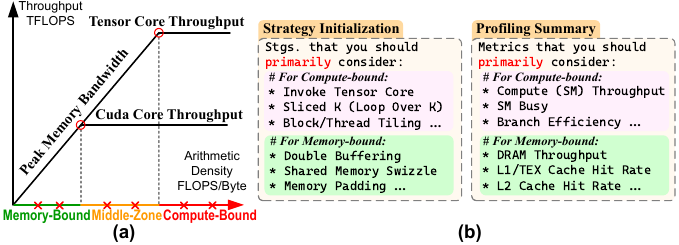}
\Description{nothing}
\caption{(a) Kernels positioned in GPU's roofline model. (b) Prompting examples for compute/memory-bound kernels. }\label{roofline}
\end{figure}

\subsection{Strategy-Coordinated Evolution Algorithm}\label{SCE}
Algorithm~\ref{SCE-Algorithm} elaborates on the SCE algorithm.
First, the population is initialized at the strategy level according to the initial kernel and $NS_0$ strategies, corresponding to part \ding{182} in Figure~\ref{framework}. 
Second, optimized kernels are generated by the Strategy Translators based on the strategies, and subsequently revised by the Kernel Revisor for syntax, function, and performance profiling (\ding{183}, \ding{184}, \ding{185}).
If a single \textit{StrategyApplication} fails to achieve significant performance improvement, the maximum optimization times can be increased to enable repeated kernel optimization along the same strategy.
Third, after reaching the maximum optimization times, the StrategyAlignment is performed to update the strategies with the optimized kernels, thereby yielding a parent population with improved performance across strategies.
Then, for evolutionary convergence, elitism preservation and tournament selection are utilized to select parent kernels and strategies, which are then crossed at the strategy-level to produce the initial population for the subsequent generation (parts \ding{187}, \ding{188}). 
Finally, after completion of multiple generations within one epoch, the top kernels are selected as initial kernels for the next epoch, and the process is repeated multiple times (part \ding{189}) for successive kernel evolution. 

The core of the SCE is to decompose the kernel crossover into a strategy-coordinated two-step process: strategy-level crossover (lines 8–11) and strategy-kernel translation (lines 5–7). 
By introducing the intermediate strategy semantics, strategy crossover is decoupled from the kernel implementation. This approach increases the success rate of kernel crossover and guides the LLM to generate kernels with more complex optimization strategies.

\subsection{Roofline-guided Prompting}\label{roofline_section}

Kernel optimization is a highly complex task that encompasses numerous optimization strategies. 
Manual kernel optimization requires consulting substantial information, such as algorithm characteristics, data layouts, and profiling reports, to guide improvements step by step. 
To prevent the LLM's optimization from focusing on unnecessary reflections and wasting effort on ineffective directions, cuPilot employs the target GPU's roofline model to transform the singular goal of kernel performance into specific objectives of different hardware utilizations.

Specifically, as the roofline model shown in Figure~\ref{roofline} (a), based on the task’s data precision, kernels located to the left of the intersection point between the throughput of the lowest-performance hardware unit (e.g., CUDA cores) and the peak memory bandwidth are identified as \textbf{memory-bound}. 
Kernels to the right of the intersection point between throughput of the highest-performance hardware unit (e.g., tensor cores) and the peak memory bandwidth are identified as \textbf{compute-bound}. 
The region between these two kernels is regarded as an \textbf{middle-zone} requiring both computation and memory optimizations.
The key insight is that computational optimizations should be prioritized for compute-bound kernels, and vice versa for memory-bound kernels.
As the prompting examples shown in Figure~\ref{roofline} (b), for compute-bound kernels, the agents are guided to generate computation optimization strategies and prioritize improving computation units utilization, with a focus on metrics like SM Throughput, Branch Efficiency, and so on.
For memory-bound kernels, emphasis is placed on enhancing the utilization of memory bandwidth, with particular attention paid to memory throughput during profiling analysis. 
Furthermore, when selecting parent individuals using a tournament selection strategy, the corresponding hardware utilization also serves as a criterion.

\begin{figure}[!t]
\centering
\includegraphics[width=3in]{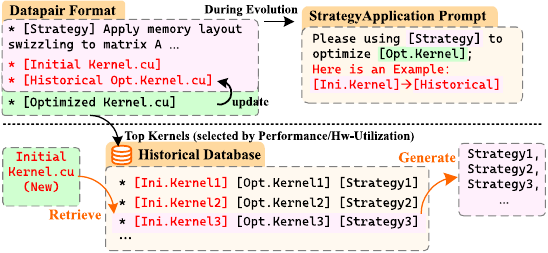}
\Description{nothing}
\caption{Historical data pair format for \textit{StrategyApplication} prompt and database construction for RAG.}\label{RAG}
\end{figure}

\subsection{Strategy-level Population Initialization}\label{RAG_section}
Population initialization defines the initial design space for kernel optimization. Coarse population initialization of previous agentic works may yield low-quality or insufficient strategies, which may constrain the upper limit of optimization effectiveness. 
To learn strategies from historical evolution, a historical database is constructed, with RAG for prompt generation.

Specifically, as illustrated in Figure~\ref{RAG}, the initial kernel, optimized kernel, and corresponding performance metrics are preserved as components of individuals and iteratively updated throughout the SCE.
On one hand, during the \textit{StrategyApplication} of Strategy Translator, these short-term historical records serve as supplementary examples, as the top of Figure~\ref{RAG} shows. On the other hand, at the end of each generation, top kernels' data pairs are archived into the RAG database. 
For subsequent strategy generation of a new kernel, the similarity between the new initial kernel and historical initial kernels in the database is retrieved, and the associated strategies are employed as prompts to facilitate the initialization. This process, known as RAG, leverages the historical optimizations to bootstrap the optimization and evolve the agentic framework.
By integrating external and historical knowledge, cuPilot achieves maximally diverse population initialization at the strategy level, thereby enhancing the overall evolutionary convergence. 

\begin{figure}[!t]
\centering
\includegraphics[width=3.3in]{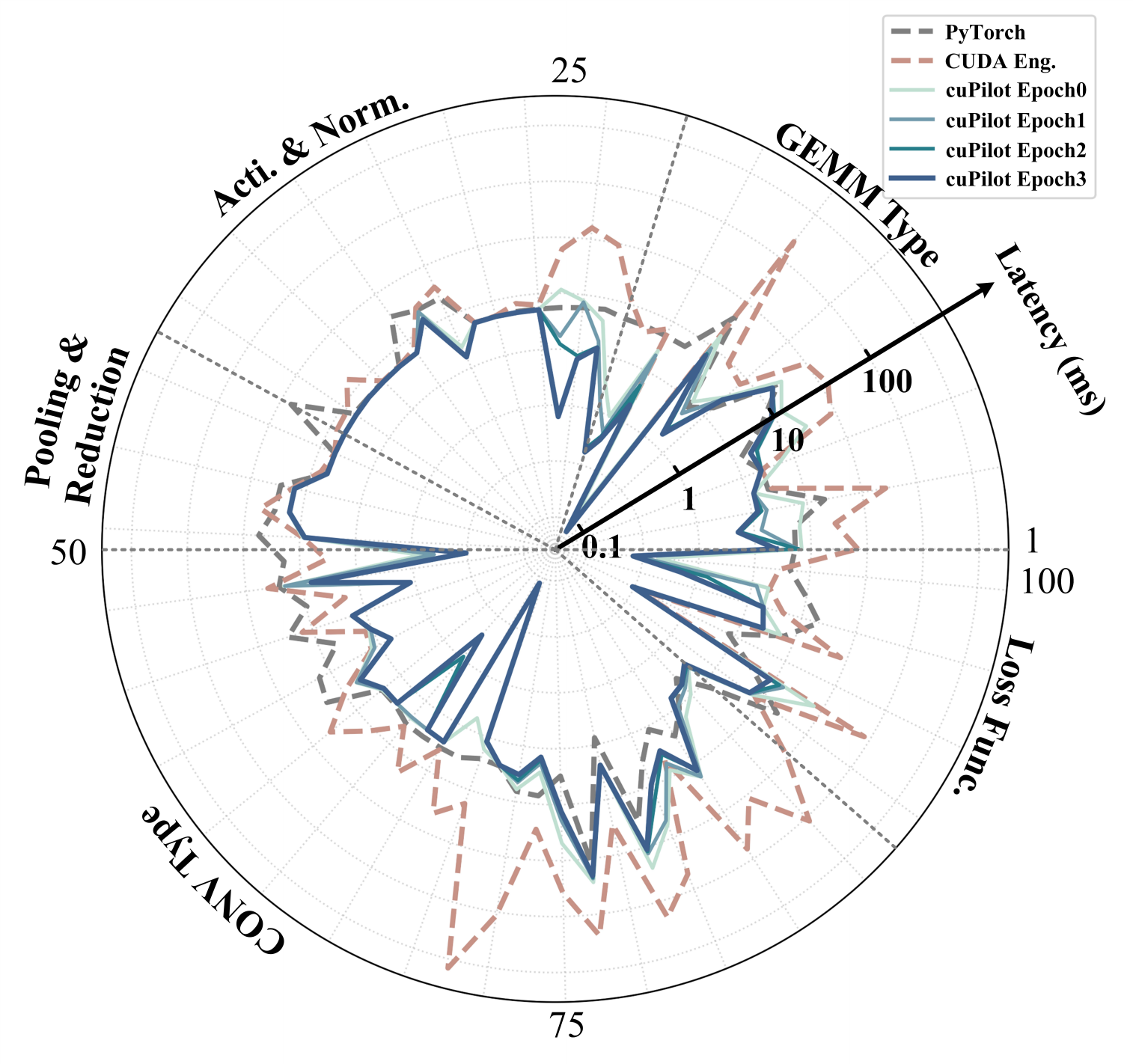}
\Description{nothing}
\caption{Performance comparison of cuPilot, PyTorch, and AI CUDA Engineer on Kernelbench benchmark.}\label{polar_fig_case}
\end{figure}

\section{Experiment}

\subsection{Experiment Setup}
The cuPilot framework is implemented and evaluated on a server equipped with eight A100 GPUs. The experiments are conducted using CUDA version 12.4. 
For kernel performance evaluation, the testing methodology from KernelBench~\cite{kernelbench} is adopted. Each individual kernel undergoes 10 warm-up runs prior to measurement, followed by 50 timed executions with different random inputs. The average latency is reported as the kernel’s performance. 
The Nsight Compute CLI. (NCU) tool is adopted for kernel profiling.
To mitigate the impact of thermal variation, the GPU frequency is locked at 1410 MHz throughout the testing process. Note that the performance of our primary baseline kernels, released by AI CUDA Engineer~\cite{sakanaAI}, was also measured under identical hardware conditions.
The overall framework is implemented in Python 3.10. To manage concurrent requests to LLMs, the asyncio library is used. 
For models, we choose the Deepseek-R1~\cite{deepseek-r1} and Gemini-2.5-pro~\cite{gemini-2.5}, as they represent the SOTA ability of thinking and reasoning. 

Firstly, to validate cuPilot's capability on kernel optimization, an evaluation is conducted on 100 kernels of KernelBench level-1. Resulting kernels are compared against both PyTorch~\cite{pytorch} and kernels released by AI CUDA Engineer~\cite{sakanaAI_hf_datasets}. 
To ensure a fair comparison, Deepseek-R1 is employed, as it is also used by AI CUDA Engineer.
Subsequently, ablation studies is performed on four representative kernels to analyze the contribution of the proposed roofline-guided prompting and RAG-enhanced population initialization. 
To validate migratability across different models, the experiment is conducted using both Deepseek and Gemini.
Finally, to better demonstrate the detailed kernel optimizations, GEMM kernels, characterized by their deep optimization potential and broad applicability, are selected as a case study. 
For a fair comparison with the AI CUDA Engineer-generated GEMMs, again, the Deepseek-R1 is employed.

\begin{figure}[!t]
\centering
\includegraphics[width=3.3in]{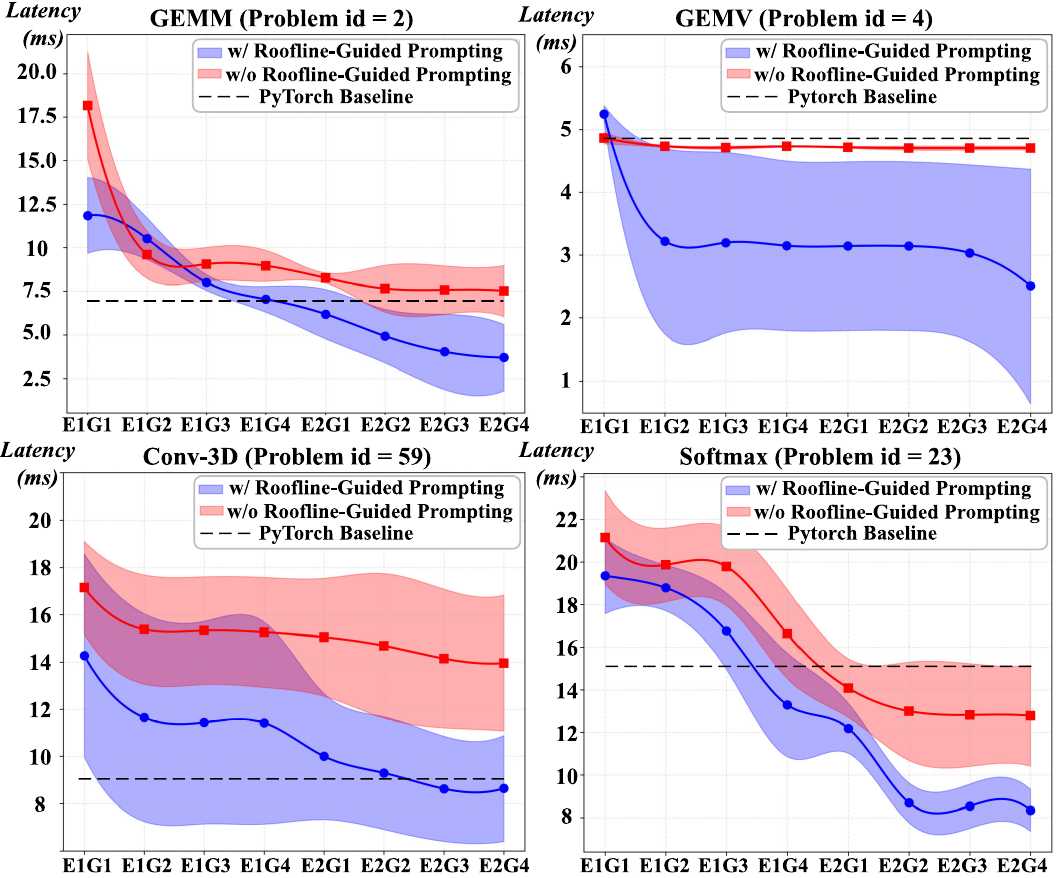}
\Description{nothing}
\caption{Ablation results for roofline-guide prompting on four representative kernels.}\label{exp_roofline}
\end{figure}

\subsection{Evaluation on KernelBench}
As shown in the Figure~\ref{polar_fig_case}, which uses a logarithmic scale, cuPilot achieves performance comparable to PyTorch's highly optimized kernels, and even delivers speedups by several times in some cases. This advantage becomes even more pronounced when compared to AI CUDA Engineer. Notably, these results of cuPilot were produced with less than 3 hours of per-kernel runtime for each epoch.

The evaluation begins with the GEMM kernels. KernelBench provides multiple GEMM kernels, covering various matrix shapes and specialized matrix types (e.g., symmetric or upper/lower triangular matrix). In our experiments, cuPilot achieves up to an average $4.06\times$ speedup over PyTorch in GEMM kernels. This improvement can be attributed to the three optimizations proposed in this work, which enable the LLM to identify bottlenecks and establish relations between high-level strategies and their low-level code implementation. Consequently, the evolutionary algorithm can effectively perform crossover, producing complex kernels that incorporate multiple optimizations. The detailed mechanism enabling these optimizations in cuPilot is further analyzed in the case study (See Section~\ref{section:case-study}). Similarly, for CONV kernels, cuPilot achieves a $1.18\times$ speedup over PyTorch. This performance gain stems from cuPilot's ability to achieve complex optimization strategies, including tiling and double buffering, as well as padding and memory swizzling to reduce bank conflicts. As for other kernels, cuPilot delivers more substantial speedups, achieving performance gains of $4.16\times$ for activation/normalization, $3.85\times$ for pooling, and $6.87\times$ for loss functions. Overall, an average speedup of $3.09\times$ is achieved.

\begin{figure}[!t]
\centering
\includegraphics[width=3.3in]{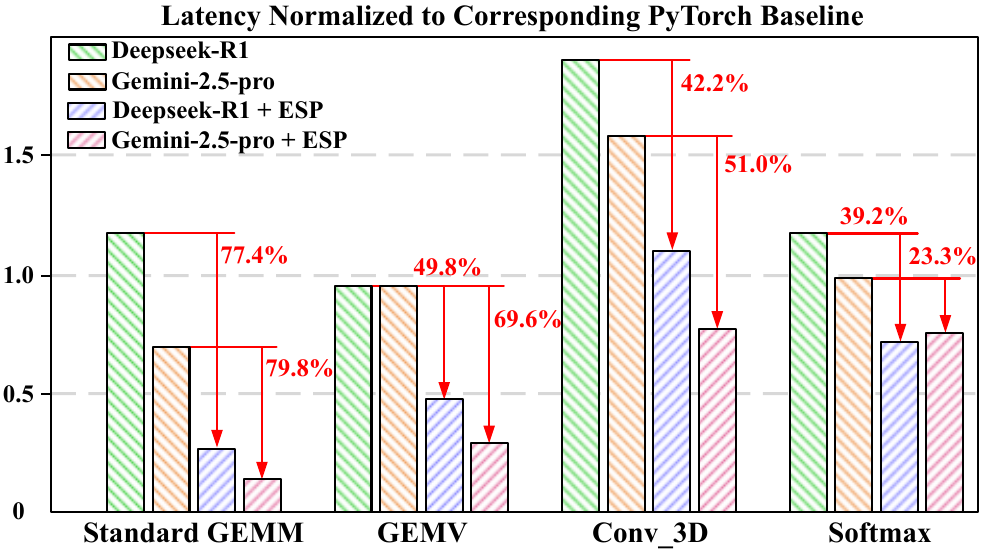}
\Description{nothing}
\caption{Ablation results for strategy-level population initialization. ESP is for External Strategy Pool.}\label{exp_RAG}
\end{figure}

\subsection{Ablation Experiments}
\textbf{(1) For roofline-guided prompting}, four representative kernels from KernelBench are evaluated. These include two compute-bound kernels: Standard GEMM and Standard 3D Convolution, and two memory-bound kernels: GEMV and Softmax. 
Evolutionary experiments are conducted using both DeepSeek-R1 and Gemini-2.5 Pro, with each configuration repeated three times. 
Each experiment is run for two epochs, with a population size of 50.
Figure~\ref{exp_roofline} presents the mean and standard deviation of the results. It is evident that kernels employing roofline-guided prompting achieve a faster reduction in latency, observed in both compute-bound and memory-bound kernels. On average, kernel latency is reduced by 44.2\% after two epochs.
The figure also includes the performance of two baselines, demonstrating that even without roofline guidance, cuPilot can achieve better performance than the AI CUDA Engineer.
\newline 
\textbf{(2) For strategy-level population initialization}, the same four representative kernels are selected, and comparative experiments are conducted with and without an external strategy pool. The control group relies solely on the LLM’s inherent kernel optimization capability to generate initial strategies. 
Figure~\ref{exp_RAG} presents a comparison of the kernel's latency after a single epoch of four generations. The results indicate that, with RAG, both DeepSeek-R1 and Gemini-2.5-Pro are able to produce more effective evolutionary strategies, leading to lower latency. 
On average, the latency across the four tasks is reduced by 54.1\% after the first epoch. 

\subsection{Case Study: GEMM Kernels}\label{section:case-study}

\begin{figure}[!t]
\centering
\includegraphics[width=3.3in]{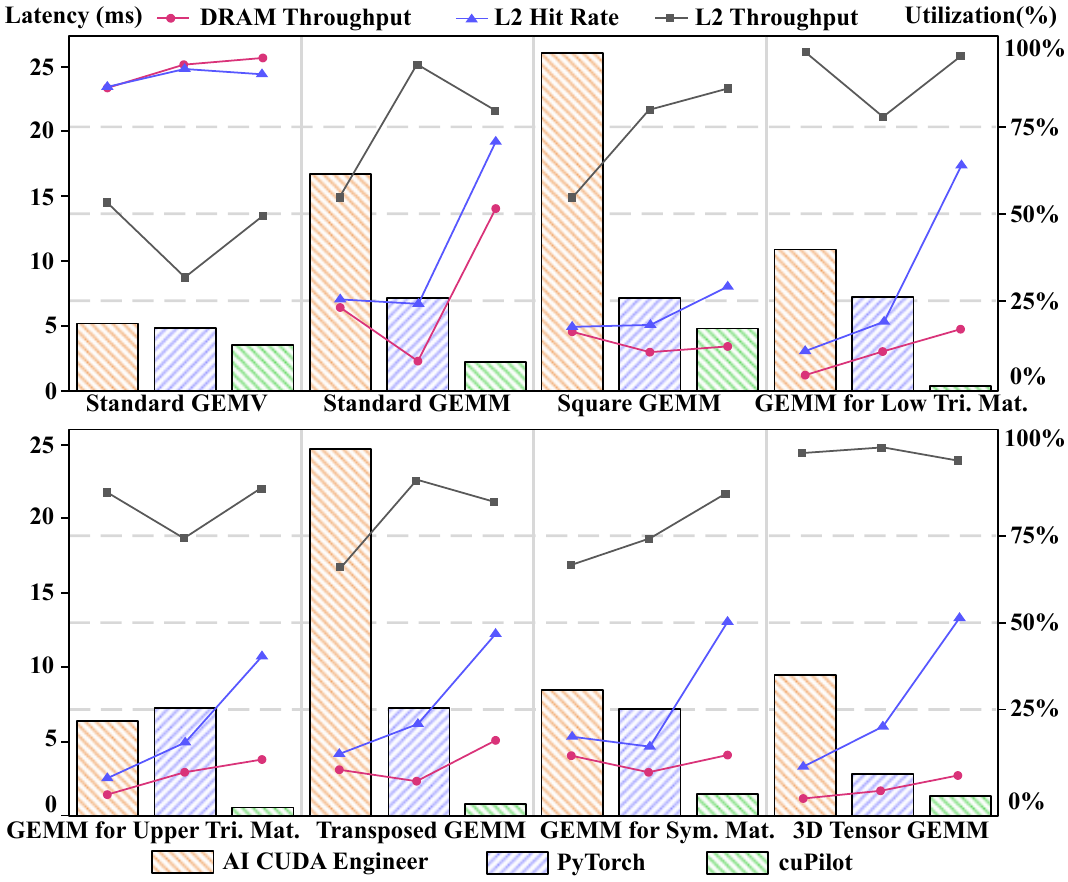}
\Description{nothing}
\caption{Performance and memory utilization comparison of GEMM kernels from different frameworks.}\label{case_study}
\end{figure}

The self-attention in Transformers relies on GEMM to compute the Q/K/V computation, making GEMM a major bottleneck in modern AI and also a primary target for optimization. However, we observe that existing works suffer from illusory optimization. For instance, AI CUDA Engineer's GEMM kernel merely acts as a wrapper for the cublas library, which bypasses the core development challenge. This case study analyzes the strategies of cuPilot's GEMM kernels to demonstrate its authentic optimization capability.

We have analyzed and compared the strategies employed in all GEMM kernels generated by cuPilot and AI CUDA Engineer, as presented in the Table~\ref{tab:case-study}. In contrast to AI CUDA Engineer, which primarily relies on basic strategies such as tiling and vectorized access, cuPilot demonstrates a superior ability to apply more sophisticated optimizations. Specifically, in computation, cuPilot leverages Tensor Cores for acceleration~\cite{tensor-dac}, while in memory access, it employs techniques like padding and layout swizzling to reduce bank conflicts, along with thread block swizzling to improve L2 cache locality~\cite{cache-dac} and hit rates. Furthermore, cuPilot optimizes the overall execution pipeline by implementing double buffering or multi-stage strategies to increase throughput further. Most notably, it exhibits the potential for fine-tuning using PTX assembly instructions~\cite{ptx-dac}, underscoring its comprehensive code generation capability.

We further validated it by profiling tools. As shown in Figure~\ref{case_study}, we compared latency and memory throughput of the three frameworks across eight GEMM cases. Compared to PyTorch, cuPilot not only maximizes computational resource utilization but also enhances memory access efficiency. The application of swizzling and asynchronous copy techniques improves L2 throughput utilization, while the use of a multi-stage pipeline further increases the throughput of DRAM and L2 cache hit rate. In contrast, AI CUDA Engineer achieves limited gains from vectorized access, and its overall memory utilization is further limited by a preliminary tiling strategy, causing it to underperform PyTorch.

\section{Conclusion}

\begin{table}[!t]
    \centering
    \fontsize{10}{6}\selectfont
    \begin{threeparttable}
    \caption{Applied strategies comparison of GEMM kernels generated by cuPilot and CUDA Engineer.}\label{tab:case-study}
    \renewcommand{\arraystretch}{2}
    \begin{tabular}{rcccc}
    \toprule
    \multirow{2}{*}{\bf Optimization Strategies} & \multicolumn{2}{c}{\bf Frameworks} \\ \cmidrule(lr){2-3}
    & This Work & CUDA Eng. \\
    \midrule
    Invoking Tensor Core & {\color[RGB]{77,153,0}{ \usym{2714}} (14/16)$^1$} & {\color[RGB]{204,0,0}{ \usym{2717}} (0/16)}  \\
    Tiling Technologies & {\color[RGB]{77,153,0}{ \usym{2714}} (15/16)} & {\color[RGB]{77,153,0}{ \usym{2714}} (14/16)}  \\
     Vectorized Access & {\color[RGB]{77,153,0}{ \usym{2714}} (10/16)} & {\color[RGB]{77,153,0}{ \usym{2714}} (2/16)}  \\
    Memory Padding & {\color[RGB]{77,153,0}{ \usym{2714}} (12/16)} & {\color[RGB]{77,153,0}{ \usym{2714}} (1/16)}  \\
    Layout/Thread Block Swizzle & {\color[RGB]{77,153,0}{ \usym{2714}} (5/16)} & {\color[RGB]{204,0,0}{ \usym{2717}} (0/16)}  \\
    Double Buffering & {\color[RGB]{77,153,0}{ \usym{2714}} (8/16)} & {\color[RGB]{77,153,0}{ \usym{2714}} (1/16)}  \\
    Multi-Stage Pipeline & {\color[RGB]{77,153,0}{ \usym{2714}} (2/16)} & {\color[RGB]{204,0,0}{ \usym{2717}} (0/16)}  \\
    Asynchronous Copy & {\color[RGB]{77,153,0}{ \usym{2714}} (5/16)} & {\color[RGB]{204,0,0}{ \usym{2717}} (0/16)}  \\
    PTX-Level Optimization & {\color[RGB]{77,153,0}{ \usym{2714}} (3/16)} & {\color[RGB]{204,0,0}{ \usym{2717}} (0/16)}  \\
    \bottomrule
    \end{tabular}
    \begin{tablenotes}
    \footnotesize
    \item[1] The Kernelbench Benchmark includes 16 GEMM cases, covering various types such as Square GEMM, Transposed GEMM, Irregular GEMM, and so on. The notation (14/16) indicates that 14 out of the 16 kernels generated for these cases incorporate the corresponding strategy.
    \end{tablenotes}
    \end{threeparttable}
\end{table}

cuPilot addresses the limitations of existing kernel evolution frameworks by introducing strategy as an intermediate semantic representation, effectively resolving the mismatches of crossover, fitness function and population initialization.
Through the SCE algorithm, roofline-guided prompting, and strategy-level population initialization, cuPilot enables LLMs to generate highly optimized CUDA kernels with complex strategies. Experimental results demonstrate an average speedup of 3.09$\times$  over PyTorch on a 100 kernels benchmark. The detailed case study on GEMM tasks showcases advanced hardware utilization and optimization strategies. 

\begin{acks}
We thank Tsing Micro and the National Center of Technology Innovation for EDA for providing experimental facilities and resources.
\end{acks}

\bibliographystyle{unsrt}
\bibliography{reference}

\end{document}